# An In-Depth Look at Information Fusion Rules and the Unification of Fusion Theories

Dr. Florentin Smarandache
The University of New Mexico
200 College Road
Gallup, NM 87301, USA
smarand@unm.edu
www.gallup.unm.edu/~smarandache/DSmT.htm

**Abstract:**
  This presentation may look like a glossary of the fusion rules and we also introduce new ones presenting their formulas and examples: Conjunctive, Disjunctive, Exclusive Disjunctive, Mixed Conjunctive-Disjunctive rules, Conditional rule, Dempster's, Yager's, Smets' TBM rule, Dubois-Prade's, Dezert-Smarandache classical and hybrid rules, Murphy's average rule, Inagaki-Lefevre-Colot-Vannoorenberghe Unified Combination rules [and, as particular cases: Iganaki's parameterized rule, Weighting Average Operator, minC (M. Daniel), and newly Proportional Conflict Redistribution rules (Smarandache-Dezert) among which PCR5 is the most exact way of redistribution of the conflicting mass to non-empty sets following the path of the conjunctive rule], Zhang's Center Combination rule, Convolutive x-Averaging, Consensus Operator (Josang), Cautious Rule (Smets), α-junctions rules (Smets), etc. and three new T-norm & T-conorm rules adjusted from fuzzy and neutrosophic sets to information fusion (Tchamova-Smarandache). Introducing the degree of union and degree of inclusion with respect to the cardinal of sets not with the fuzzy set point of view, besides that of intersection, many fusion rules can be improved.
There are corner cases where each rule might have difficulties working or may not get an expected result.
   As a conclusion, since no theory neither rule fully satisfy all needed applications, the author proposes a Unification of Fusion Theories extending the power and hyper-power sets from previous theories to a Boolean algebra obtained by the closures of the frame of discernment under union, intersection, and complement of sets (for non-exclusive elements one considers a fuzzy or neutrosophic complement, i.e. a function which is not involutive).
And, at each application, one selects the most appropriate model, rule, and algorithm of implementation.

**Download this paper** from **arXiv** at:
http://xxx.lanl.gov/ftp/cs/papers/0410/0410033.pdf .

**ACM:** I.2.3. (Artificial Intelligence: Uncertainty, fuzzy/neutrosophic and probabilistic reasoning)

**Introduction.**
Let's consider the frame of discernment $\Theta = \{\theta_1, \theta_2, \ldots, \theta_n\}$, with $n \geq 2$, and two sources of information:
$m_1(.), m_2(.): S^\Theta \to [0, 1]$.

For the simplest frame $\Theta = \{\theta_1, \theta_2\}$ one can define a mass matrix as follows:

|        | $\theta_1$ | $\theta_2$ | $\theta_1 \cup \theta_2$ | $\theta_1 \cap \theta_2$ | $\mathscr{C}\theta_1$ | $\mathscr{C}\theta_2$ | $\mathscr{C}(\theta_1 \cap \theta_2)$ | $\phi$ |
|--------|-----|-----|-----|-----|-----|-----|-----|-----|
| $m_1(.)$ | $m_{11}$ | $m_{12}$ | $m_{13}$ | $m_{14}$ | $m_{15}$ | $m_{16}$ | $m_{17}$ | $m_{18}$ |
| $m_2(.)$ | $m_{21}$ | $m_{22}$ | $m_{23}$ | $m_{24}$ | $m_{25}$ | $m_{26}$ | $m_{27}$ | $m_{28}$ |

In calculations we take in account only the focal elements, i.e. those for which $m_1(.)$ or $m_2(.) > 0$. In the Shafer's model one has only the first three columns of the mass matrix, corresponding to $\theta_1, \theta_2, \theta_1 \cup \theta_2$, while in the Dezert-Smarandache free model only the first four columns corresponding to $\theta_1, \theta_2, \theta_1 \cup \theta_2, \theta_1 \cap \theta_2$. But here we took the general case in order to include the possible complements (negations) as well.
We note the combination of these bbas, using any of the below rule "r", by
$m_r = m_1 \otimes_r m_2$.

All the rules below are extended from their power set $2^\Theta = (\Theta, \cup) = \{\phi, \theta_1, \theta_2, \theta_1 \cup \theta_2\}$, which is a set closed under union, or hyper-power set $D^\Theta = (\Theta, \cup, \cap) = \{\phi, \theta_1, \theta_2, \theta_1 \cup \theta_2, \theta_1 \cap \theta_2\}$ which is a distributive lattice called hyper-power set, to the super-power set $S^\Theta = (\Theta, \cup, \cap, \mathscr{C}) = \{\phi, \theta_1, \theta_2, \theta_1 \cup \theta_2, \theta_1 \cap \theta_2, \mathscr{C}\theta_1, \mathscr{C}\theta_2, \mathscr{C}(\theta_1 \cap \theta_2)\}$, which is a Boolean algebra with respect to the union, intersection, and complement ($\mathscr{C}$ is the complement).
Of course, all of these can be generalized for $\Theta$ of dimension $n \geq 2$ and for any number of sources $s \geq 2$.

Similarly one defines the mass matrix, power-set, hyper-power set, and super-power set for the general frame of discernment.

**Two types of uncertainties:**

1) **aleatory** (also called stochastic, variability, irreducible, or uncertainty of type A), which arises because the system behave in many ways;
2) and **epistemic** (also called uncertainty of type B, or subjective, state of knowledge, reducible), which is caused by the lack of knowledge.

- Information fusion deals with combination of epistemic uncertainty and paradoxist/conflicting information.
- About 32 combination fusion rules have been collected, old and new introduced; some connected with corresponding Fusion Theories.
- Each rule works better in some conditions and less in others; more rules required for an application.

**Algebraic properties of fusion rules:**
- *Commutativity*

If S1 and S2 are two sensors/bbas and $\sigma$ a combination rule, then $\sigma(S1,S2)=\sigma(S2,S1)$.
- *Associativity*

If S1,S2, and S3 are three sensors/bbas and $\sigma$ a combination rule, then
$\sigma(\sigma(S1,S2),S3)=\sigma(S1,\sigma(S2,S3))$.
Some rules are *quasi-associative*: to preserve associativity the conjunctive rule's result (it is based on) is stored in computer, and then combined with the new evidence.
- *Idempotence*

If S1 is a sensor/bba and $\sigma$ a combination rule, then $\sigma(S1,S1)=S1$.
- *Continuity*

If S1 gives values close to S1', then $\sigma(S1,S2)$ should give values close to $\sigma(S1',S2)$.
- *Vacuum belief assignment* (VBA), which is VBA(total ignorance)=1, should act as a neutral element for $\sigma$.
- *Markovian process*: $\sigma(S1,S2,VBA)= \sigma(S1,S2)$.

Some rules are *quasi-Markovian*.

Here it is a list of all rules we could collect from various sources:

1. **Conjunctive Rule**:

If both sources of information are telling the truth, then we apply the conjunctive rule, which means consensus between them (or their common part):

$\forall A \in S^\Theta$, one has $m_\cap(A) = \sum_{\substack{X_1,X_2 \in S^\wedge \Theta \\ X_1 \cap X_2 = A}} m_1(X_1)m_2(X_2)$.

where the **Conflicting Mass** is:

$k_{12} = m_\cap(\phi) = \sum_{\substack{X_1,X_2 \in S^\wedge \Theta \\ X_1 \cap X_2 = \phi}} m_1(X_1)m_2(X_2)$.

Weakness: If a belief is 0, no matter what others are, the combination is 0.

2. **Disjunctive Rule** (Dubois-Prade, 1986):

If at least one source of information is telling the truth, we use the optimistic disjunctive rule:

$m_\cup(\phi) = 0$, and $\forall A \in S^\Theta \setminus \phi$, one has $m_\cup(A) = \sum_{\substack{X_1,X_2 \in S^\wedge \Theta \\ X_1 \cup X_2 = A}} m_1(X_1)m_2(X_2)$.

Weakness: Similarly if a belief is 0, no matter what others are, the combination is 0.

3. **Exclusive Disjunctive Rule** (Dubois-Prade, 1986):

If only one source of information is telling the truth, but we don't know which one, then one uses the exclusive disjunctive rule based on the fact that $X_1 \veebar X_2$ means either $X_1$ is true, or $X_2$ is true, but not both in the same time (in set theory let's use $X_1 \,e\!\cup X_2$ for exclusive disjunctive):

$m_{e\cup}(\phi) = 0$, and $\forall A \in S^\Theta \setminus \phi$, one has $m_{e\cup}(A) = \sum_{\substack{X_1, X_2 \in S^{\wedge\Theta} \\ X_{1e} \cup X_2 = A}} m_1(X_1) m_2(X_2)$.

**4. Mixed Conjunctive Disjunctive Rule** (Dubois-Prade, 1986):
This is a mixture of the previous three rules in any possible way.

As an example, suppose we have four sources of information and we know that: either the first two are telling the truth or the third, or the fourth is telling the truth.
The mixed formula becomes:
$m_{\cap\cup}(\phi) = 0$, and $\forall A \in S^\Theta \setminus \phi$, one has $m_{\cap\cup}(A) = \sum_{\substack{X_1, X_2, X_3, X_4 \in S^{\wedge\Theta} \\ ((X_1 \cap X_2) \cup X_3)e \cup X_4 = A}} m_1(X_1) m_2(X_2) m_3(X_3) m_4(X_4)$.

**5. Conditional Rule**:
This rule is considered in any fusion theory, and it looks like the conditional probability but it is different. We use the conditional rule when there exists a bba, say $m_c(.)$, such that for an hypothesis, say A, one has $m_c(A) = 1$ (i.e. when the subjective certainty of an hypothesis to occur is given by an expert). Then we simply combine this $m_c(.)$ with another given bba, using whatever rule of combination is given in that fusion theory.

**6. Dempster's Rule**:
This is the most used fusion rule in applications and this rule influenced the development of other rules. Shafer (1976) has developed the Dempster-Shafer Theory of Evidence based on the model that all hypotheses in the frame of discernment are exclusive and the frame is exhaustive.

$m_D(\phi) = 0$, and $\forall A \in S^\Theta \setminus \phi$, one has $m_D(A) = \dfrac{1}{1 - k_{12}} \cdot \sum_{\substack{X_1, X_2 \in S^{\wedge\Theta} \\ X_1 \cap X_2 = A}} m_1(X_1) m_2(X_2)$

Zadeh's Example Generalized:

|       | A   | B   | C |
|-------|-----|-----|---|
| S1    | 1-e | 0   | e |
| S2    | 0   | 1-e | e |
| S1&S2 | 0   | 0   | 1 |

where e>0 is a tiny number.
Wrong result! Belief on C is 1 whatever the value of e is!

Dezert-Smarandache-Khoshnevisan Example:

|      | A   | B  | C  | D  |
|------|-----|----|----|----|
| S1   | .6  | 0  | .4 | 0  |
| S2   | 0   | .7 | 0  | .3 |
| S1&S2| 0/0 | 0/0| 0/0| 0/0|

Conflict is 1.
The rule doesn't work at all!

## 7. Murphy's Statistical Average Rule:

If we consider that the bbas are important from a statistical point of view, then one averages them:
$\forall A \in S^\Theta$, one has $m_M(A) = \frac{1}{2}[m_1(A) + m_2(A)]$.

Or, more general, $m_{mixing}(A) = \frac{1}{2}[w_1 m_1(A) + w_1 m_2(A)]$, where $w_1$, $w_2$ are weights reflecting the reliability of sources.

|           | A   | B   | C   | A∪B |
|-----------|-----|-----|-----|-----|
| $m_1$     | .2  | .4  | .3  | .1  |
| $m_2$     | .1  | .3  | .4  | .2  |
| $m_1$&$m_2$ | .15 | .35 | .35 | .15 |

Weakness: Too mechanical, not well justified.

## 8. Dezert-Smarandache Classic Rule:

This is a generalization of the conjunctive rule from the power set to the hyper-power set.
$\forall A \in S^\Theta$, one has $m_{DSmC}(A) = \sum_{\substack{X_1, X_2 \in S^\Theta \\ X_1 \cap X_2 = A}} m_1(X_1) m_2(X_2)$.

It can also be extended on the Boolean algebra $(\Theta, \cup, \cap, \mathcal{C})$ in order to include the complements (or negations) of elements.

Weakness: If a belief is 0, no matter what others are the result is 0.

## 9. Dezert-Smarandache Hybrid Rule:

It is an extension of the Dubois-Prade rule for the dynamic fusion. The middle sum in the below formula does not occur in Dubois-Prade's rule, and it helps in the transfer of the masses of empty sets - whose disjunctive forms are also empty - to the total ignorance.

$m_{DSmH}(\phi) = 0$, and $\forall A \in S^\Theta \setminus \phi$ one has

$$m_{DSmH}(A) = \sum_{\substack{X_1,X_2 \in S^\wedge\Theta \\ X_1 \cap X_2 = A}} m_1(X_1)m_2(X_2) + \sum_{\substack{X_1,X_2 \in \phi \\ (A=U) \vee \{U \in \phi \wedge A=I\}}} m_1(X_1)m_2(X_2) + \sum_{\substack{X_1,X_2 \in S^\wedge\Theta \\ U(c\{X_1 \cap X_2\})=A \\ X_1 \cap X_2 \in \phi}} m_1(X_1)m_2(X_2)$$

where U is the disjunctive form of $X_1 \cup X_2$ and it is defined as follows:
$U(X) = X$ if X is a singleton, $U(X_1 \cap X_2) = U(X_1) \cup U(X_2)$, and $U(X_1 \cup X_2) = U(X_1) \cup U(X_2)$;
while $I = \theta_1 \cup \theta_2 \cup \ldots \cup \theta_n$ is the total ignorance, and $c\{X_1 \cap X_2\}$ is the canonical form of this expression, i.e. the simplest form [for example, $c\{(A \cap B) \cap (A \cup B \cup C)\} = A \cap B$].
Formally the canonical form has the properties:
i) $c(\phi) = \phi$;
ii) if A is a singleton, then $c(A) = A$;
iii) if $A \subseteq B$, then $c(A \cap B) = A$ and $c(A \cup B) = B$;
iiii) the second and third properties apply for any number of sets.

|         | A   | B   | C   | A∩B= {.3<x<.4} | A∪C | B∪C |
|---------|-----|-----|-----|----------------|-----|-----|
| $m_1$   | .5  | .2  | .3  |                |     |     |
| $m_2$   | .4  | .4  | .2  |                |     |     |
| $m_1\&m_2$ | .20 | .08 | .06 | .28            | .22 | .16 |

Suppose the targets' cross sections are A={x<.4}, B={.3<x<.6}, C={x>.8}.
DSm hybrid model (one intersection A&B=nonempty ).
This example proves the necessity of allowing intersections of elements in the frame of discernment.   [Shafer's model doesn't apply here.]
Dezert-Smarandache Theory of Uncertain and Paradoxist Reasoning (DSmT) is the only one which accepts intersections of elements.

Weakness: Too complex to compute.

### 10. Smets' TBM Rule:
Smets does not transfer the conflicting mass, but keeps it on the empty set, meaning that $m(\phi) > 0$ signifies that there might exist other hypotheses we don't know of in the frame of discernment (this is called an *open world*).

$$m_S(\phi) = k_{12} = \sum_{\substack{X_1,X_2 \in S^\wedge\Theta \\ X_1 \cap X_2 = \phi}} m_1(X_1)m_2(X_2).$$

and $\forall A \in S^\Theta \backslash \phi$, one has $m_S(A) = \sum_{\substack{X_1,X_2 \in S^\wedge\Theta \\ X_1 \cap X_2 = A}} m_1(X_1)m_2(X_2)$.

Weakness: At high conflict or in dynamic fusion one gets less information (much mass assigned to the empty set)!  Empty set interpreted as the set of missing hypotheses.

### 11. Yager's Rule:
Yager transfers the conflicting mass to the total ignorance.

$m_Y(\phi) = 0$, $m_Y(I) = m_1(I)m_2(I) + \sum_{\substack{X_1, X_2 \in S^{\wedge\Theta} \\ X_1 \cap X_2 = \phi}} m_1(X_1)m_2(X_2)$ where I = total ignorance, and $\forall A \in S^\Theta \setminus \{\phi, I\}$, one has $m_Y(A) = \sum_{\substack{X_1, X_2 \in S^{\wedge\Theta} \\ X_1 \cap X_2 = A}} m_1(X_1)m_2(X_2)$.

Weakness: It is the least specific result (total ignorance becomes big fast)!

## 12. Dubois-Prade's Rule:

This rule is based on the principle that of two sources are in conflict, then at least one is true, and thus transfers the conflicting mass $m(A \cap B) > 0$ to $A \cup B$.

$m_{DP}(\phi) = 0$,

and $\forall A \in S^\Theta \setminus \phi$ one has

$m_{DP}(A) = \sum_{\substack{X_1, X_2 \in S^{\wedge\Theta} \\ X_1 \cap X_2 = A}} m_1(X_1)m_2(X_2) + \sum_{\substack{X_1, X_2 \in S^{\wedge\Theta} \\ X_1 \cup X_2 = A \\ X_1 \cap X_2 = \phi}} m_1(X_1)m_2(X_2)$.

Incomplete result:

|          | A   | B   | C=empty | A∪B |
|----------|-----|-----|---------|-----|
| $m_1$    | .2  | .4  | .3      | .1  |
| $m_2$    | .1  | .3  | .4      | .2  |
| $m_1\&m_2$ | .28 | .48 | 0       | .12 |

At time t one finds out that target C is not the real (C=empty).
Weakness: In dynamic fusion if a singleton becomes empty the rule provides incomplete results! [sum of masses is .88 < 1]

## 13. Weighted Operator (Unification of the Rules) by Iganaki-Lefevre-Colot-Vannoorenberghe, 1991, 2002):

$m_{WO}(\phi) = w_m(\phi) \cdot k_{12}$,

and $\forall A \in S^\Theta \setminus \phi$, one has $m_{WO}(A) = \sum_{\substack{X_1, X_2 \in S^{\wedge\Theta} \\ X_1 \cap X_2 = A}} m_1(X_1)m_2(X_2) + w_m(A) \cdot k_{12}$

where $w_m(A) \in [0, 1]$ for any $A \in S^\Theta$ and $\sum_{X \in S^{\wedge\Theta}} w_m(X) = 1$ and $w_m(A)$ are called weighting factors.

## 14. Inagaki's Unified Parameterized Combination Rule (1991):

$\forall A \in S^\Theta \setminus \{\phi, I\}$, one has $m_p^U(A) = [1 + p \cdot k_{12}] \sum_{\substack{X_1, X_2 \in S^{\wedge\Theta} \\ X_1 \cap X_2 = A}} m_1(X_1)m_2(X_2)$,

and $m_p^U(\phi) = 0$, $m_p^U(I) = [1+p \cdot k_{12}] \sum_{\substack{X_1, X_2 \in S^\wedge \Theta \\ X_1 \cap X_2 = I}} m_1(X_1)m_2(X_2) + [1+p \cdot k_{12}-p]k_{12}$

where the parameter $0 \leq p \leq 1 / [1-k_{12}-m_\cap(I)]$, and $k_{12}$ is the conflict.
The determination of parameter p, used for normalization, is not well justified in the literature, but may be found through experimental data, simulations, expectations (Tanaka-Klir, 1999). The greater is the parameter p, the greater is the change to the evidence.

**15. The Weighted Average Operator (WAO)** (Jøsang-Daniel-Vannoorenberghe, 2003):
This rule consists in first, applying the conjunctive rule to the bbas $m_1(.)$ and $m_2(.)$ and second, redistribute the total conflicting mass $k_{12}$ to all nonempty sets in $S^\Theta$ proportionally with their mass averages, i.e. for the set, say A, proportionally with the weighting factor:
$w_{JDV}(A, m_1, m_2) = \frac{1}{2} (m_1(A) + m_2(A))$.
The authors do not give an analytical formula for it.
WAO does not work in degenerated cases:

|  | A | B(t) | C | $k_{12}$=conflict |
|---|---|---|---|---|
| $m_1$ | .2 | .4 | .3 | |
| $m_2$ | .1 | .3 | .4 | |
| $m_1$&$m_2$(t) ConjRule | .02 | .12 | .12 | .74 |
| $m_1$&$m_2$(t+1) ConjRule | .02 | B(t+1)=empty 0 | .12 | .74+.12=.86 |
| $m_1$&$m_2$ WAO | .149 | 0 | .421 | 0 |

In dynamic fusion at time t+1 one finds out that target B is not real (B=empty).

Weakness: In degenerate cases doesn't work (sum of masses .570 < 1 in Shafer's model).

**15.1.** Smarandache-Dezert (2004) independently developed a **Proportional Conflict Redistribution Rule (PCR1)**, which similarly consists in first, applying the conjunctive rule to the bbas $m_1(.)$ and $m_2(.)$ and second, redistribute the total conflicting mass $k_{12}$ to all nonempty sets in $S^\Theta$ proportionally with their nonzero mass sum, i.e. for the set, say A, proportionally with the weighting factor:
$w_{SD}(A, m_1, m_2) = m_1(A) + m_2(A) \neq 0$.
P. Smets pointed out that PCR1 gives the same result as the WAO for nondegenerated cases.

The analytical formula for PCR1, nondegenerated and degenerated cases, is:
$m_{PCR1}(\phi) = 0$, and

$$\forall A \in S^\Theta \setminus \phi, \text{ one has } m_{PCR1}(A) = \sum_{\substack{X_1, X_2 \in S^\wedge \Theta \\ X_1 \cap X_2 = A}} m_1(X_1) m_2(X_2) + \frac{c_{12}(A)}{d_{12}} \cdot k_{12},$$

where $c_{12}(A)$ is the sum of masses corresponding to the set A, i.e. $c_{12}(A) = m_1(A) + m_2(A) \neq 0$, $d_{12}$ is the sum of nonzero masses of all nonempty sets in $S^\Theta$ assigned by the sources $m_1(.)$ and $m_2(.)$ [in many cases $d_{12} = 2$, but in degenerated cases it can be less], and $k_{12}$ is the total conflicting mass.

PCR1 extends WAO, since PCR1 works also for the degenerate cases when all column sums of all non-empty sets are zero because in such cases, the conflicting mass is transferred to the non-empty disjunctive form of all non-empty sets together; when this disjunctive form happens to be empty, then one can consider an open world (i.e. the frame of discernment might contain new hypotheses) and thus all conflicting mass is transferred to the empty set.

The VBA (vacuous belief assignment) is the bba $m_v$(total ignorance) = 1.
For the cases of the combination of only one non-vacuous belief assignment $m_1(.)$ with the vacuous belief assignment $m_v(.)$ where $m_1(.)$ has mass assigned to an empty element, say $m_1(.) > 0$ as in Smets' TBM, or as in DSmT dynamic fusion where one finds out that a previous non-empty element A, whose mass $m_1(A) > 0$, becomes empty after a certain time, then this mass of an empty set has to be transferred to other elements using PCR1, but for such case $[m_1 \otimes m_v](.)]$ is different from $m_1(.)$. This severe draw-back of WAO and PCR1 forces us to develop the next PCR rules satisfying the neutrality property of VBA with better redistributions of the conflicting information.

|  | A | B(t) | C | $k_{12}$=conflict |
|---|---|---|---|---|
| $m_1$ | .2 | .4 | .3 |  |
| $m_2$ | .1 | .3 | .4 |  |
| $m_1 \& m_2$(t) ConjRule | .02 | .12 | .12 | .74 |
| $m_1 \& m_2$(t+1) ConjRule | .02 | B(t+1)=empty 0 | .12 | .74+.12=.86 |
| $m_1 \& m_2$ WAO | .149 | 0 | .421 | 0 |
| $m_1 \& m_2$ PCR1 | .278 | 0 | .722 | 0 |

WAO example solved with PCR1 (sum of masses is 1).

Weakness: Not all subsets deserve to receive part of the conflicting mass.

**15.2.** Smarandache-Dezert (2004) then developed more improved versions of **Proportional Conflict Redistribution Rule (PCR2-4)**:
In the **PCR2**, the total conflicting mass $k_{12}$ is redistributed only to the non-empty sets involved in the conflict (not to all non-empty sets as in WAO and PCR1) proportionally with respect to their corresponding non-empty column sum in the mass matrix. The redistribution is then more exact (accurate) than in PCR1 and WAO. A nice feature of PCR2 is the preservation of the neutral impact of the VBA and of course its ability to deal with all cases/models.

$m_{PCR2}(\phi) = 0$, and
$\forall A \in S^\Theta \setminus \phi$ and A involved in the conflict, one has

$$m_{PCR2}(A) = \sum_{\substack{X_1, X_2 \in S^\wedge \Theta \\ X_1 \cap X_2 = A}} m_1(X_1)m_2(X_2) + \frac{c_{12}(A)}{e_{12}} \cdot k_{12},$$

while for a set $B \in S^\Theta \setminus \phi$ not involved in the conflict one has:

$$m_{PCR2}(B) = \sum_{\substack{X_1, X_2 \in S^\wedge \Theta \\ X_1 \cap X_2 = B}} m_1(X_1)m_2(X_2),$$

where $c_{12}(A)$ is the non-zero sum of the column of X in the mass matrix, i.e. $c_{12}(A) = m_1(A) + m_2(A) \neq 0$, $k_{12}$ is the total conflicting mass, and $e_{12}$ is the sum of all non-zero column sums of all non-empty sets only involved in the conflict (in many cases $e_{12} = 2$, but in some degenerate cases it can be less). In the degenerate case when all column sums of all non-empty sets involved in the conflict are zero, then the conflicting mass is transferred to the non-empty disjunctive form of all sets together which were involved in the conflict. But if this disjunctive form happens to be empty, then one considers an open world (i.e. the frame of discernment might contain new hypotheses) and thus all conflicting mass is transferred to the empty set.

A non-empty set $X \in S^\Theta$ is considered involved in the conflict if there exists another set $Y \in S^\Theta$ such that $X \cap Y = 0$ and $m_{12}(X \cap Y) > 0$. This definition can be generalized for $s \geq 2$ sources.

**15.3. PCR3** transfers partial conflicting masses, instead of the total conflicting mass. If an intersection is empty, say $A \cap B = \phi$, then the mass $m(A \cap B) > 0$ of the partial conflict is transferred to the non-empty sets A and B proportionally with respect to the non-zero sum of masses assigned to A and respectively to B by the bbas $m_1(.)$ and $m_2(.)$. The PCR3 rule works if at least one set between A and B is non-empty and its column sum is non-zero. When both sets A and B are empty, or both corresponding column sums of the mass matrix are zero, or only one set is non-empty and its column sum is zero, then the mass $m(A \cap B)$ is transferred to the non-empty disjunctive form $u(A) \cup u(B)$ [which is defined as follows: $u(A) = A$ if A is a singleton, $u(A \cap B) = u(A \cup B) = u(A) \cup u(B)$]; if this disjunctive form is empty then $m(A \cap B)$ is transferred to the non-empty total ignorance in a closed world approach or to the empty set if one prefers to adopt the Smets' open world approach; but if even the total ignorance is empty (a completely de generate case) then one considers an open world (i.e. new hypotheses might be in the frame of discernment) and the conflicting mass is transferred to the empty set, which means that the original problem has no solution in the close world initially chosen for the problem.

$m_{PCR3}(\phi) = 0$, and $\forall A \in S^{\Theta}\setminus\phi$ one has

$$m_{PCR3}(A) = \sum_{\substack{X_1, X_2 \in S^{\wedge}\Theta \\ X_1 \cap X_2 = A}} m_1(X_1)m_2(X_2) + c_{12}(A) \cdot \sum_{\substack{X \in S^{\wedge}\Theta \setminus A \\ X \cap A = \phi}} \frac{m_1(A)m_2(X) + m_2(A)m_1(X)}{c_{12}(A) + c_{12}(X)} +$$

$$\sum_{\substack{X_1, X_2 \in S^{\wedge}\Theta \setminus A \\ X_1 \cap X_2 = \phi \\ u(X_1) \cup u(X_2) = A}} [m_1(X_1)m_2(X_2) + m_1(X_2)m_2(X_1)] +$$

$$\Psi_{\Theta}(A) \cdot \sum_{\substack{X_1, X_2 \in S^{\wedge}\Theta \setminus A \\ X_1 \cap X_2 = \phi \\ u(X_1) = u(X_2) = A}} [m_1(X_1)m_2(X_2) + m_2(X_1)m_1(X_2)]$$

where $c_{12}(A)$ is the non-zero sum of the mass matrix column corresponding to the set A, and the total ignorance characteristic function $\Psi_{\Theta}(A) = 1$ if A is the total ignorance, and 0 otherwise.

**15.4. PCR4** improves Milan Daniel's below minC rule. After applying the conjunctive rule, Daniel uses the proportionalization with respect to the results of the conjunctive rule, and not with respect to the masses assigned to each nonempty set by the sources of information as done in PCR1-3 or the next PCR5.
PCR4 also uses the proportionalization with respect to the results of the conjunctive rule, but with PCR4 the conflicting mass $m_{12}(A \cap B) > 0$ when $A \cap B = \phi$ is distributed to A and B only because only A and B were involved in the conflict {$A \cup B$ was not involved in the conflict since $m_{12}(A \cap B) = m_1(A)m_2(B) + m_2(A)m_1(B)$}, while minC [both its versions a) and b)] redistributes the conflicting mass $m_{12}(A \cap B)$ to A, B, and $A \cup B$. Also, for the mixed sets such as $C \cap (A \cup B) = \phi$ the conflicting mass $m_{12}(C \cap (A \cup B)) > 0$ is distributed to C and $A \cup B$ because only them were involved in the conflict by PCR4, while minC version a) redistributes $m_{12}(C \cap (A \cup B))$ to C, $A \cup B$, $C \cup A \cup B$ and minC version b) redistributes $m_{12}(C \cap (A \cup B))$ even worse to A, B, C, $A \cup B$, $A \cup C$, $B \cup C$, $A \cup B \cup C$.

$m_{PCR4}(\phi) = 0$, and $\forall A \in S^{\Theta}\setminus\phi$ one has

$$m_{PCR4}(A) = m_{12}(A) + \sum_{\substack{X \in S^{\wedge}\Theta \setminus A \\ X \cap A = \phi}} m_{12}(X) \frac{m_{12}(A \cap X)}{m_{12}(A) + m_{12}(X)}$$

where $m_{12}(.)$ is the conjunctive rule, and all denominators $m_{12}(A) + m_{12}(X) \neq 0$; (if a denominator corresponding to some X is zero, the fraction it belongs to is discarded and the mass $m_{12}(A \cap X)$ is transferred to A and X using PCR3.

minC Example solved with PCR4:

|  | A | B ∪ C | A ∪ B ∪ C | A ∩ (B ∪ C) |
|---|---|---|---|---|
| $m_1$ | .5 | .1 | .4 | |
| $m_2$ | .7 | .2 | .1 | |
| $m_1$&$m_2$ ConjRule | .68 | .11 | .04 | .17 |
| $m_1$&$m_2$ minC a), b) PCR4 | .819277 .826329 | .132530 .133671 | .048193 .04 | 0 0 |

Redistribution better in PCR4 since A∪B∪C not involved in conflict doesn't deserve extra-mass.

Weakness: A more exact redistribution can be done (see PCR5).

**14.5.** PCR5 is the most mathematically exact form of redistribution of the conflicting mass to non-empty sets which follows backwards the tracks of the conjunctive rule formula. But it is the most difficult to implement. In order to better understand it, let's start with some examples:

Example 14.5.1.
Suppose one has the mass matrix
     A     B     A∪B
$m_1$   0.6   0     0.4
$m_2$   0     0.3   0.7
The conjunctive rule yields:
$m_{12}$   0.42   0.12   0.28
and the conflicting mass $k_{12} = 0.18$.
Only A and B were involved in the conflict,
$k_{12} = m_{12}(A \cap B) = m_1(A)m_2(B) + m_2(A)m_1(B) = m_1(A)m_2(B) = 0.6 \cdot 0.3 = 0.18$.
Therefore, 0.18 should be distributed to A and B proportionally with respect to 0.6 and 0.3 {i.e. the masses assigned to A and B by the sources $m_1(.)$ and $m_2(.)$} respectively. Let x be the conflicting mass to be redistributed to A and y the conflicting mass to be redistributed to B (out of 0.18), then:
$$\frac{x}{0.6} = \frac{y}{0.3} = \frac{x+y}{0.6+0.3} = \frac{0.18}{0.9} = 0.2,$$
whence $x = 0.6 \cdot 0.2 = 0.12$, $y = 0.3 \cdot 0.2 = 0.06$, which is normal since 0.6 is twice bigger than 0.3. Thus:
$m_{PCR4}(A) = 0.42 + 0.12 = 0.54$,
$m_{PCR4}(B) = 0.12 + 0.06 = 0.18$,
$m_{PCR4}(A \cup B) = 0.28 + 0 = 0.28$.
This result is the same as PCR2-3.

Example 14.5.2.
Let's modify a little the previous example and have the mass matrix
     A     B     A∪B
$m_1$   0.6   0     0.4

m₂   0.2   0.3   0.5
The conjunctive rule yields:
m₁₂   0.50   0.12   0.20
and the conflicting mass k₁₂ = 0.18.
The conflict k₁₂ is the same as in previous example, which means that m₂(A) = 0.2 did not have any impact on the conflict; why?, because m₁(B) = 0.
A and B were involved in the conflict, A∪B is not, hence only A and B deserve a part of the conflict, A∪B does not deserve.
With PCR5 one redistributes the conflicting mass 0.18 to A and B proportionally with the masses m₁(A) and m₂(B) respectively, i.e. identically as above. The mass m₂(A) = 0.2 is not considered to the weighting factors of redistribution since it did not increase or decrease the conflicting mass. One obtains x = 0.12 and y = 0.06, which added to the previous masses yields:
m_PCR4(A) = 0.50 + 0.12 = 0.62,
m_PCR4(B) = 0.12 + 0.06 = 0.18,
m_PCR4(A∪B) = 0.20.
This result is different from all PCR1-4.

Example 14.5.3.
Let's modify a little the previous example and have the mass matrix
           A     B     A∪B
m₁       0.6   0.3   0.1
m₂       0.2   0.3   0.5
The conjunctive rule yields:
m₁₂   0.44   0.27   0.05
and the conflicting mass
k₁₂ = m₁₂(A∩B) = m₁(A)m₂(B) + m₂(A)m₁(B) = 0.6·0.3 + 0.2·0.3 = 0.18 + 0.06 = 0.24.
Now the conflict is different from the previous two examples, because m₂(A) and m₁(B) are both non-null. Then the partial conflict 0.18 should be redistributed to A and B proportionally to 0.6 and 0.3 respectively (as done in previous examples, and we got x1 = 0.12 and y1 = 0.06), while 0.06 should be redistributed to A and B proportionally to 0.2 and 0.3 respectively.
For the second redistribution one similarly calculate the proportions:
$$\frac{x2}{0.2} = \frac{y2}{0.3} = \frac{x2+y2}{0.2+0.3} = \frac{0.06}{0.5} = 0.12,$$
whence x = 0.2·0.12 = 0.024, y = 0.3·0.12 = 0.036. Thus:
m_PCR4(A) = 0.44 + 0.12 + 0.024 = 0.584,
m_PCR4(B) = 0.27 + 0.06 + 0.036 = 0.366,
m_PCR4(A∪B) = 0.05 + 0 = 0.050.
This result is different from PCR1-4.

Formula:
$m_{PCR5}(\phi) = 0$, and $\forall A \in S^\Theta \setminus \phi$ one has
$$m_{PCR5}(A) = m_{12}(A) + \sum_{\substack{X \in S^\Theta \\ X \cap A = \phi}} [\frac{m1(A)^2 \cdot m2(X)}{m1(A)+m2(X)} + \frac{m2(A)^2 \cdot m1(X)}{m2(A)+m1(X)}],$$

where $m_{12}(.)$ is the conjunctive rule, and all denominators are different from zero; if a denominator is zero, the fraction it belongs to is discarded.

**16. minC (minimum conflict) Rule** (M. Daniel, 2000):
This rule improves Dempster's rule since the distribution of the conflicting mass is done from each partial conflicting mass to the subsets of the sets involved in partial conflict proportionally with respect to the results of the conjunctive rule for each such subset. It goes by types of conflicts. The author does not provide an analytical formula for this rule. minC is commutative, associative, and non-idempotent.
Let $m_{12}(X \cap Y) > 0$ be a conflicting mass, where $X \cap Y = \phi$, and X, Y may be singletons or mixed sets (i.e. unions or intersections of both of singletons).
minC has two versions, minC a) and minC b), which differs from the way the redistribution is done: either to the subsets X, Y, and X∪Y in version a), or to all subsets of P(u(X), u(Y)) in version b).
One applies the conjunctive rule, and then the partial conflict, say $m_{12}(A \cap B)$, when $A \cap B = \phi$, is redistributed to A, B, A∪B proportionally to the masses $m_{12}(A)$, $m_{12}(B)$, and $m_{12}(A \cup B)$ respectively in both versions a) and b). PCR4 redistributes the conflicting mass to A and B since only them were involved in the conflict.
But for a mixed set, as shown above, say $C \cap (A \cup B) = \phi$, the conflicting mass $m_{12}(C \cap (A \cup B)) > 0$ is distributed by PCR4 to C and A∪B because only them were involved in the conflict, while the minC version a) redistributes $m_{12}(C \cap (A \cup B))$ to C, A∪B, C∪A∪B, and minC version b) redistributes $m_{12}(C \cap (A \cup B))$ even worse to A, B, C, A∪B, A∪C, B∪C, A∪B∪C.
Another example is that the mass $m_{12}(A \cap B \cap C)) > 0$, when $A \cap B \cap C = \phi$, is redistributed in both versions minC a) and minC b) to A, B, C, A∪B, A∪C, B∪C, A∪B∪C.
When the conjunctive rule results are zero for all the nonempty sets that are redistributed conflicting masses, the conflicting mass is averaged to each such set.

|  | A | B ∪ C | A ∪ B ∪ C | A ∩ (B ∪ C) |
|---|---|---|---|---|
| $m_1$ | .5 | .1 | .4 |  |
| $m_2$ | .7 | .2 | .1 |  |
| $m_1 \& m_2$ ConjRule | .68 | .11 | .04 | .17 |
| $m_1 \& m_2$ minC a), b) | .819277 | .132530 | .048193 | 0 |

Weakness: Not all subsets deserve to receive part of the conflicting mass. Here minC coincides with Dempster's.

**17. Consensus Operator (CO)** (Jøsang, 2001):
It is defined only on binary frames of discernment. CO doesn't work on non-exclusive elements (i.e. on models with nonempty intersections of sets).

On the frame $\Theta = \{\theta_1, \theta_2\}$ of exclusive elements, $\theta_2$ is considered the complement/negation of $\theta_1$.

If the frame of discernment has more than two elements, then by a simple or normal coarsening it is possible to derive a binary frame containing any element A and its complement $\mathcal{C}(A)$. Let m(.) be a bba on a (coarsened) frame $\Theta = \{A, \mathcal{C}(A)\}$, then one defines an opinion resulted from this bba is:
$w_A = (b_A, d_A, u_A, \alpha_A)$, where $b_A = m(A)$ is the belief of A, $d_A = m(\mathcal{C}(A))$ is the disbelief of A, $u_A = m(A \cup \mathcal{C}(A))$ is the uncertainty of A, and $\alpha_A$ represents the atomicity of A. Of course $b_A + d_A + u_A = 1$, for $A \neq \phi$.

The relative atomicity expresses information about the relative size of the state space (i.e. the frame of discernment). For every operator, the relative atomicity of the output belief is computed as a function of the input belief operands. The relative atomicity of the input operands is determined by the state space circumstances, or by a previous operation in case that operation's output is used as input operand. The relative atomicity itself can also be uncertain, and that's what's called state space uncertainty. Possibly the state space uncertainty is a neglected problem in belief theory. It relates to Smet's open world, and to DSm paradoxical world. In fact, the open world breaks with the "exhaustive" assumption, and the paradoxical world breaks with the "exclusive" assumption of classic belief theory. CO is commutative, associative, and non-idempotent.

Having two experts with opinions on the same element A,
$w_{1A} = (b_{1A}, d_{1A}, u_{1A}, \alpha_{1A})$ and
$w_{2A} = (b_{2A}, d_{2A}, u_{2A}, \alpha_{2A})$, one first computes
$k = u_{1A} + u_{2A} - u_{1A} \cdot u_{2A}$.
Let's note by $w_{12A} = (b_{12A}, d_{12A}, u_{12A}, \alpha_{12A})$ the consensus opinion between $w_{1A}$ and $w_{2A}$. Then:
  a) for $k \neq 0$ one has:
  $b_{12A} = (b_{1A} \cdot u_{2A} + b_{2A} \cdot u_{1A}) / k$
  $d_{12A} = (d_{1A} \cdot u_{2A} + d_{2A} \cdot u_{1A}) / k$
  $u_{12A} = (u_{1A} \cdot u_{2A}) / k$
  $$\alpha_{12A} = \frac{\alpha_{1A} u_{2A} + \alpha_{2A} u_{1A} - (\alpha_{1A} + \alpha_{2A}) u_{1A} u_{2A}}{u_{1A} + u_{2A} - 2 u_{1A} u_{2A}}$$
  b) for $k = 0$ one has:
  $b_{12A} = (\gamma_{12A} \cdot b_{1A} + b_{2A}) / (\gamma_{12A} + 1)$
  $d_{12A} = (\gamma_{12A} \cdot d_{1A} + d_{2A}) / (\gamma_{12A} + 1)$
  $u_{12A} = 0$
  $\alpha_{12A} = (\gamma_{12A} \cdot \alpha_{1A} + \alpha_{2A}) / (\gamma_{12A} + 1)$
  where $\gamma_{12A} = u_{2A} / u_{1A}$ represents the relative dogmatism between opinions $w_{1A}$ and $w_{2A}$.

The formulas are not justified, and there is not a well-defined method for computing the relative atomicity of an element when a bba is known.

For frames of discernment of size greater than n, or with many sources, or in the open world it is hard to implement CO.

A bba m(.) is called *Bayesian* on the frame $\Theta = \{\theta_1, \theta_2\}$ of exclusive elements if $m(\theta_1 \cup \theta_2) = 0$, otherwise it is called *non Bayesian*.

If one bba is Bayesian, say $m_1(.)$, and another is not, say $m_2(.)$, then the non Bayesian bba is ignored! See below $m_{CO}(.) = m_1(.)$:

Example

|       | A   | B   | A∪B |
|-------|-----|-----|-----|
| $m_1$ | 0.3 | 0.7 | 0.0 |
| $m_2$ | 0.8 | 0.1 | 0.1 |
| $m_{CO}$ | 0.3 | 0.7 | 0.0 |

Because

|          | $b_A$ | $d_A$ | $u_A$ | $\alpha_A$ |
|----------|-------|-------|-------|------------|
| $m_{1A}$ | 0.3   | 0.7   | 0.0   | 0.5        |
| $m_{2A}$ | 0.8   | 0.1   | 0.1   | 0.5        |

$\alpha_{1A} = \alpha_{2A} = \dfrac{|A \cap \Theta|}{|\Theta|} = 0.5$, where $|X|$ means the cardinal of X, whence $\alpha_{12A} = 0.5$.

Similarly one computes the opinion on B, because:

|          | $b_B$ | $d_B$ | $u_B$ | $\alpha_B$ |
|----------|-------|-------|-------|------------|
| $m_{1B}$ | 0.7   | 0.3   | 0.0   | 0.5        |
| $m_{2B}$ | 0.1   | 0.8   | 0.1   | 0.5        |

If both bbas are Bayseian, then one uses their arithmetic mean.

### 18. Zhang's Center Combination Rule (1994):

$\forall A \in S^{\Theta}$, one has $m_Z(A) = k \cdot \sum\limits_{\substack{X_1, X_2 \in S^{\wedge}\Theta \\ X_1 \cap X_2 = A}} \dfrac{|X_1 \cap X_2|}{|X_1| \cdot |X_2|} m_1(X_1) m_2(X_2)$.

where k is a renormalization factor, $|X|$ is the cardinal of the set X, and

$r(X_1, X_2) = \dfrac{|X_1 \cap X_2|}{|X_1| \cdot |X_2|}$ represents the degree (measure) of intersection of the sets $X_1$ and $X_2$.

In Dempster's approach the degree of intersection was assumed to be 1.

The degree of intersection could be defined in many ways, for example

$r(X_1, X_2) = \dfrac{|X_1 \cap X_2|}{|X_1 \cup X_2|}$ could be better defined this way since if the intersection is empty the degree of intersection is zero, while for the maximum intersection, i.e. when $X_1 = X_2$, the degree of intersection is 1.

One can attach the $r(X_1, X_2)$ to many fusion rules.

Weakness: The degree of intersection may be defined in many ways.

### 19. Convolutive x-Averaging (Ferson-Kreinovich, 2002):

∀ A∈S^Θ, one has $m_X(A) = \sum_{\substack{X_1,X_2 \in S^{\wedge\Theta} \\ (X_1+X_2)/2=A}} m_1(X_1)m_2(X_2)$

This rule works for hypotheses defined as subsets of the set of real numbers.

|  | A=[2,5] <br> $m_1(A)$=.6 | B=[1,3] <br> $m_1(B)$=.4 |
|---|---|---|
| A=[2,5] <br> $m_2(A)$=.7 | [2,5] <br> $m_1\&m_2([2,5])$=.42 | [1.5,4] <br> $m_1\&m_2([1.5,4])$=.28 |
| B=[1,3] <br> $m_2(B)$=.3 | [1.5,4] <br> $m_1\&m_2([1.5,4])$=.18 | [1,3] <br> $m_1\&m_2([1,3])$=.12 |

Hence, $m_1\&m_2([1.5,4])$ = .46.

**20. α-junctions Rules** (P. Smets, 1999):
Are generalizations of the above Conjunctive and Disjunctive Rules, and they are parameterized with respect to α ∈ [0, 1].
Smets finds the rules for the elementary frame of discernment Θ with two hypotheses, using a matrix operator $\mathbf{K}_X$, for each X∈{φ, A, B, A∪B} and shows that it is possible to extend them by iteration to larger frames of discernment.
These rules are more theoretical and hard to apply.

**21. Cautious Rule** (P. Smets, 2000):
This rule is just theoretical. Also, Smets does not provide a formula or a method for calculating this rule. He states this *Theorem*:
Let $m_1$, $m_2$ be two bbas, and q1, q2 their corresponding commonality functions, and $SP(m_1)$, $SP(m_2)$ the set of specializations of $m_1$ and $m_2$ respectively. Then the hyper-cautious combination rule
$m_{1⊛2}$ = min{m, m∈SP($m_1$)∩SP($m_2$)},
and the commonality of $m_{1⊛2}$ is $q_{12}$ where $q_{12}(A) = \min\{q_1(A), q_2(A)\}$.

We recall that the *commonality function* of a bba m(.) is q: $S^\Theta \to [0, 1]$ such that:
$q(A) = \sum_{\substack{X \in S^{\wedge\Theta} \\ X \supseteq A}} m(X)$ for all A ∈ $S^\Theta$.

Now a few words about the least commitment and specialization.

a) *Least Commitment, or Minimum Principle*, means to assign a missing mass of a bba or to transfer a conflicting mass to the least specific element in the frame of discernment (in most of the cases to the partial ignorances or to the total ignorance).
"The Principle of Minimal Commitment consists in selecting the least committed belief function in a set of equally justified belief functions. This selection procedure does not always lead to a unique solution in which case extra requirements are added. The principle formalizes the idea that one should never give more support than justified to any subset of Ω. It satisfies a form of skepticism, of a commitment, of conservatism in the allocation of our belief. In its spirit, it is not far from what the probabilists try to achieve with the maximum entropy principle (see Dubois and Prade 1987, Hsia, 1991, Smets1993b)." [P. Smets]

b) About *specialization* (Yager, 1986, Dubois and Prade, 1986, Kruse and Schwecke, 1990, Delgado and Moral,1987, Smets, 2000):
Suppose at time $t_o$ one has the evidence $m_0(.)$ which gives us the value of an hypothesis A as $m_0(A)$. When a new evidence $m_1(.)$ comes in at time $t_1 > t_0$, then $m_0(A)$ might flow down to the subsets of A therefore towards a more specific information. The impact of a new bba might result in a redistribution of the initial mass of A, $m_0(A)$, towards its more specific subsets. Thus $m_1(.)$ is called a specialization of $m_0(.)$.

**22-26. Other fusion rules:**
**Yen's rule** is related to fuzzy set, while the **p-boxes method** to upper and lower probabilities (neutrosophic probability is a generalization of upper and lower probability), also **Yao and Wong's Qualitative Rule** is not numerical therefore not required in engineering - we did not include them.
There is in the literature another rule, called **Baldwin's rule**, based on partial normalization, but we did not find enough information on it.
There also exists a **Besnard's rule** defined on a Lindenbaum algebra.

**27-32. Replacing the Conjunctive Rule and Disjunctive Rule with the T-norm and T-conorm versions respectively** (Tchamova-Smarandache):

These rules started from the T-norm and T-conorm respectively in fuzzy and neutrosophic logics, where the "and" logic operator $\wedge$ corresponds in fusion to the conjunctive rule, while the "or" logic operator $\vee$ corresponds to the disjunctive rule. While the logic operators deal with degrees of truth and degrees of falsehood, the fusion rules deal with degrees of belief and degrees of disbelief of hypotheses.

A **T-norm** is a function $T_n: [0, 1]^2 \to [0, 1]$, defined in fuzzy/neutrosophic set theory and fuzzy/neutrosophic logic to represent the "intersection" of two fuzzy/neutrosophic sets and the fuzzy/neutrosophic logical operator "and" respectively. Extended to the fusion theory the T-norm will be a substitute for the conjunctive rule.
The T-norm satisfies the conditions:
 a) Boundary Conditions:   $T_n(0, 0) = 0$, $T_n(x, 1) = x$.
 b) Commutativity: $T_n(x, y) = T_n(y, x)$.
 c) Monotonicity: If  $x \leq u$ and $y \leq v$, then $T_n(x, y) \leq T_n(u, v)$.
 d) Associativity: $T_n(T_n(x, y), z) = T_n(x, T_n(y, z))$.
There are many functions which satisfy the T-norm conditions. We present below the most known ones:
The Algebraic Product T-norm:
 $T_{n\text{-algebraic}}(x, y) = x \cdot y$
The Bounded T-norm:

$T_{\text{n-bounded}}(x, y) = \max\{0, x+y-1\}$

The Default (min) T-norm (introduced by Zadeh):

$T_{\text{n-min}}(x, y) = \min\{x, y\}$.

A **T-conorm** is a function $T_c: [0, 1]^2 \to [0, 1]$, defined in fuzzy/neutrosophic set theory and fuzzy/neutrosophic logic to represent the "union" of two fuzzy/neutrosophic sets and the fuzzy/neutrosophic logical operator "or" respectively. Extended to the fusion theory the T-conorm will be a substitute for the disjunctive rule.

The T-conorm satisfies the conditions:
a) Boundary Conditions: $T_c(1, 1) = 1$, $T_c(x, 0) = x$.
b) Commutativity: $T_c(x, y) = T_c(y, x)$.
c) Monotonicity: if $x \leq u$ and $y \leq v$, then $T_c(x, y) \leq T_c(u, v)$.
d) Associativity: $T_c(T_c(x, y), z) = T_c(x, T_c(y, z))$.

There are many functions which satisfy the T-conorm conditions. We present below the most known ones:

The Algebraic Product T-conorm:

$T_{\text{c-algebraic}}(x, y) = x+y-x \cdot y$

The Bounded T-conorm:

$T_{\text{c-bounded}}(x, y) = \min\{1, x+y\}$

The Default (max) T-conorm (introduced by Zadeh):

$T_{\text{c-max}}(x, y) = \max\{x, y\}$.

Then, the T-norm Fusion rule is defined as follows:

$$m_{\cap 12}(A) = \sum_{\substack{X,Y \in \Theta \\ X \cap Y = A}} Tn(m1(X), m2(Y))$$

and the T-conorm Fusion rule is defined as follows:

$$m_{\cup 12}(A) = \sum_{\substack{X,Y \in \Theta \\ X \cup Y = A}} Tc(m1(X), m2(Y))$$

The min T-norm rule yields results, very closed to Conjunctive Rule. It satisfies the principle of neutrality of the vacuous bba, reflects the majority opinion, converges towards idempotence. It is simpler to apply, but needs normalization.

What is missed it is a strong justification of the way of presenting the fusion process. But we think, the consideration between two sources of information as a vague relation, characterized with the particular way of association between focal elements, and corresponding degree of association (interaction) between them is reasonable. (Albena Tchamova)

Min rule can be interpreted as an optimistic lower bound for combination of bba and the below Max rule as a prudent/pessimistic upper bound. (Jean Dezert)

The T-norm and T-conorm are commutative, associative, isotone, and have a neutral element.

**Degree of Intersection:**

The degree of intersection measures the percentage of overlapping region of two sets $X_1$, $X_2$ with respect to the whole reunited regions of the sets using the cardinal of sets not the fuzzy set point of view:

$$d(X_1 \cap X_2) = \frac{|X_1 \cap X_2|}{|X_1 \cup X_2|},$$

where |X| means cardinal of the set X.

For the minimum intersection/overlapping, i.e. when $X_1 \cap X_2 = \phi$, the degree of intersection is 0, while for the maximum intersection/overlapping, i.e. when $X_1 = X_2$, the degree of intersection is 1.

**Degree of Union** (Smarandache, 2004)**:**

The degree of intersection measures the percentage of non-overlapping region of two sets $X_1$, $X_2$ with respect to the whole reunited regions of the sets using the cardinal of sets not the fuzzy set point of view:

$$d(X_1 \cup X_2) = \frac{|X_1 \cup X_2| - |X_1 \cap X_2|}{|X_1 \cup X_2|}.$$

For the maximum non-overlapping, i.e. when $X_1 \cap X_2 = \phi$, the degree of union is 1, while for the minimum non-overlapping, i.e. when $X_1 = X_2$, the degree of union is 1. The sum of degrees of intersection and union is 1 since they complement each other.

**Degree of Inclusion** (Smarandache, 2004)**:**

The degree of intersection measures the percentage of the included region $X_1$ with respect to the includant region $X_2$:

Let $X_1 \subseteq X_2$, then

$$d(X_1 \subseteq X_2) = \frac{|X_1|}{|X_2|}.$$

$d(\phi \subseteq X_2) = 0$ because nothing is included in $X_2$, while $d(X_2 \subseteq X_2) = 1$ because $X_2$ is fulfilled by inclusion. By definition $d(\phi \subseteq \phi) = 1$.

And we can generalize the above degree for $n \geq 2$ sets.

**Improvements of Believe and Plausible Functions** (Smarandache, 2004):

Thus the Bel(.) and Pl(.) functions can incorporate in their formulas the above degrees of inclusion and intersection respectively:

Believe function improved:

$$\forall\, A \in S^\Theta \setminus \phi \text{ one has } \mathrm{Bel}_d(A) = \sum_{\substack{X \in S^\wedge\Theta \\ X \subseteq A}} \frac{|X|}{|A|} m(X)$$

Plausible function improved:

$$\forall\, A \in S^\Theta \setminus \phi \text{ one has } \mathrm{Pl}_d(A) = \sum_{\substack{X \in S^\wedge\Theta \\ X \cap A \neq \phi}} \frac{|X \cap A|}{|X \cup A|} m(X)$$

Disjunctive rule improved:

$\forall\ A \in S^\Theta \setminus \phi$, one has $m_{\cup d}(A) = k_{\cup d} \cdot \sum\limits_{\substack{X_1, X_2 \in S^{\wedge\Theta} \\ X_1 \cup X_2 = A}} \dfrac{|X_1 \cup X_2| - |X_1 \cap X_2|}{|X_1 \cup X_2|} m_1(X_1) m_2(X_2)$,

where $k_{\cup d}$ is a constant of renormalization.

Dezert-Smarandache classical rule improved:

$\forall\ A \in S^\Theta$, one has $m_{DSmCd}(A) = k_{DSmCd} \cdot \sum\limits_{\substack{X_1, X_2 \in S^{\wedge\Theta} \\ X_1 \cap X_2 = A}} \dfrac{|X_1 \cap X_2|}{|X_1 \cup X_2|} m_1(X_1) m_2(X_2)$,

where $k_{DSmCd}$ is a constant of renormalization.
It is similar with the Zhang's Center Combination rule extended on the Boolean algebra $(\Theta, \cup, \cap, \mathcal{C})$ and using another definition for the degree of intersection.

Dezert-Smarandache hybrid rule improved:
$\forall\ A \in S^\Theta \setminus \phi$ one has

$m_{DSmHd}(A) = k_{DSmHd} \cdot \{ \sum\limits_{\substack{X_1, X_2 \in S^{\wedge\Theta} \\ X_1 \cap X_2 = A}} \dfrac{|X_1 \cap X_2|}{|X_1 \cup X_2|} m_1(X_1) m_2(X_2) + \sum\limits_{\substack{X_1, X_2 \in \phi \\ (A=U) \vee \{U \in \phi \wedge A=I\}}} m_1(X_1) m_2(X_2) +$

$\sum\limits_{\substack{X_1, X_2 \in S^{\wedge\Theta} \\ X_1 \cup X_2 = A \\ X_1 \cap X_2 = \phi}} \dfrac{|X_1 \cup X_2| - |X_1 \cap X_2|}{|X_1 \cup X_2|} m_1(X_1) m_2(X_2) \}$

where $k_{DSmHd}$ is a constant of renormalization.

Smets' rule improved:

$m_S(\phi) = k_{12} = k_{Sd} \cdot \sum\limits_{\substack{X_1, X_2 \in S^{\wedge\Theta} \\ X_1 \cap X_2 = \phi}} \dfrac{|X_1 \cap X_2|}{|X_1 \cup X_2|} m_1(X_1) m_2(X_2)$,

and $\forall\ A \in S^\Theta \setminus \phi$, one has $m_S(A) = k_{Sd} \cdot \sum\limits_{\substack{X_1, X_2 \in S^{\wedge\Theta} \\ X_1 \cap X_2 = A}} \dfrac{|X_1 \cap X_2|}{|X_1 \cup X_2|} m_1(X_1) m_2(X_2)$,

where $k_{Sd}$ is a constant of renormalization.

Yager's rule:

$m_Y(\phi) = 0$, $m_Y(I) = k_{Yd} \cdot \{ m_1(I) m_2(I) + \sum\limits_{\substack{X_1, X_2 \in S^{\wedge\Theta} \\ X_1 \cap X_2 = \phi}} \dfrac{|X_1 \cap X_2|}{|X_1 \cup X_2|} m_1(X_1) m_2(X_2) \}$

and $\forall\ A \in S^\Theta \setminus \{\phi, I\}$, one has $m_{Yd}(A) = k_{Yd} \cdot \sum\limits_{\substack{X_1, X_2 \in S^{\wedge\Theta} \\ X_1 \cap X_2 = A}} \dfrac{|X_1 \cap X_2|}{|X_1 \cup X_2|} m_1(X_1) m_2(X_2)$.

where I = total ignorance and $k_{Yd}$ is a constant of renormalization.

Dubois-Prade's rule:
$m_{DP}(\phi) = 0$,
and $\forall A \in S^\Theta \setminus \phi$ one has

$$m_{DP}(A) = k_{DPd} \cdot \{ \sum_{\substack{X_1, X_2 \in S^\wedge \Theta \\ X_1 \cap X_2 = A}} \frac{|X_1 \cap X_2|}{|X_1 \cup X_2|} m_1(X_1) m_2(X_2) +$$

$$\sum_{\substack{X_1, X_2 \in S^\wedge \Theta \\ X_1 \cup X_2 = A \\ X_1 \cap X_2 = \phi}} \frac{|X_1 \cup X_2| - |X_1 \cap X_2|}{|X_1 \cup X_2|} m_1(X_1) m_2(X_2) \},$$

where $k_{DPd}$ is a constant of renormalization.

**Unification of Fusion Theories (UFT)** (Smarandache, 2004).
As a conclusion, since no theory neither rule fully satisfy all needed applications, the author proposes a Unification of Fusion Theories extending the power and hyper-power sets from previous theories to a Boolean algebra obtained by the closures of the frame of discernment under union, intersection, and complement of sets (for non-exclusive elements one considers a fuzzy or neutrosophic complement – a function not necessarily involutive).
And, at each application, one selects the most appropriate model, rule, and algorithm of implementation.

Since everything depends on the application/problem to solve, this scenario looks like a logical chart designed by the programmer in order to write and implement a computer program, or even like a cooking recipe.

Here it is the scenario attempting for a unification and reconciliation of the fusion theories and rules:

1) If all sources of information are reliable, then apply the conjunctive rule, which means consensus between them (or their common part):
2) If some sources are reliable and others are not, but we don't know which ones are unreliable, apply the disjunctive rule as a cautious method (and no transfer or normalization is needed).
3) If only one source of information is reliable, but we don't know which one, then use the exclusive disjunctive rule based on the fact that $X_1 \veebar X_2 \veebar \ldots \veebar X_n$ means either $X_1$ is reliable, or $X_2$, or and so on, or $X_n$, but not two or more in the same time.
4) If a mixture of the previous three cases, in any possible way, use the mixed conjunctive-disjunctive rule.
5) If we know the sources which are unreliable, we discount them. But if all sources are fully unreliable (100%), then the fusion result becomes the vacuum bba (i.e. $m(\Theta) = 1$, and the problem is indeterminate. We need to get new sources which are reliable or at least they are not fully unreliable.
6) If all sources are reliable, or the unreliable sources have been discounted (in the <u>default case</u>), then use the DSm classic rule (which is commutative, associative, Markovian) on Boolean algebra $(\Theta, \cup, \cap, \mathcal{C})$, no matter what contradictions (or model) the problem has.

I emphasize that the super-power set $S^\Theta$ generated by this Boolean algebra contains singletons, unions, intersections, and complements of sets.

7) If the sources are considered from a statistical point of view, use *Murphy's average rule* (and no transfer or normalization is needed).

8) In the case the model is not known (the <u>default case</u>), it is prudent/cautious to use the free model (i.e. all intersections between the elements of the frame of discernment are non-empty) and DSm classic rule on $S^\Theta$, and later if the model is found out (i.e. the constraints of empty intersections become known), one can adjust the conflicting mass at any time/moment using the DSm hybrid rule.

9) Now suppose the model becomes known [i.e. we find out about the contradictions (= empty intersections) or consensus (= non-empty intersections) of the problem/application]. Then:

    9.1) If an intersection A∩B is not empty, we keep the mass m(A∩B) on A∩B, which means consensus (common part) between the two hypotheses A and B (i.e. both hypotheses A and B are right) [here one gets *DSmT*].

    9.2) If the intersection A∩B = $\phi$ is empty, meaning contradiction, we do the following:

        9.2.1) if one knows that between these two hypotheses A and B one is right and the other is false, but we don't know which one, then one transfers the mass m(A∩B) to m(A∪B), since A∪B means at least one is right [here one gets *Yager's* if n=2, or *Dubois-Prade,* or *DSmT*];

        9.2.2) if one knows that between these two hypotheses A and B one is right and the other is false, and we know which one is right, say hypothesis A is right and B is false, then one transfers the whole mass m(A∩B) to hypothesis A (nothing is transferred to B);

        9.2.3) if we don't know much about them, but one has an optimistic view on hypotheses A and B, then one transfers the conflicting mass m(A∩B) to A and B (the nearest specific sets in the Specificity Chains) [using *Dempster's*, *PCR2-5*]

        9.2.4) if we don't know much about them, but one has a pessimistic view on hypotheses A and B, then one transfers the conflicting mass m(A∩B) to A∪B (the more pessimistic the further one gets in the Specificity Chains: (A∩B) ⊂ A ⊂ (A∪B) ⊂ I); this is also the <u>default case</u> [using *DP's*, *DSm hybrid rule*, *Yager's*];
if one has a very pessimistic view on hypotheses A and B then one transfers the conflicting mass m(A∩B) to the total ignorance in a closed world [*Yager's*, *DSmT*], or to the empty set in an open world [*TBM*];

        9.2.5.1) if one considers that no hypothesis between A and B is right, then one transfers the mass m(A∩B) to other non-empty sets (in the case more hypotheses do exist in the frame of discernment) - different from A, B, A∪B - for the reason that: if A and B are not right then there is a bigger chance that other hypotheses in the frame of discernment have a higher subjective probability to occur; we do this transfer in a **closed world** [DSm hybrid rule]; but, if it is an **open world**, we can transfer the mass m(A∩B) to the empty set leaving room for new possible hypotheses [here one gets *TBM*];

        9.2.5.2) if one considers that none of the hypotheses A, B is right and no other hypothesis exists in the frame of discernment (i.e. n = 2 is the size of the frame of

discernment), then one considers the **open world** and one transfers the mass to the empty set [here *DSmT* and *TBM* converge to each other].

Of course, this procedure is extended for any intersections of two or more sets: A∩B∩C, etc. and even for mixed sets: A∩ (B∪C), etc.

If it is a dynamic fusion in a real time and associativity and/or Markovian process are needed, use an algorithm which transforms a rule (which is based on the conjunctive rule and the transfer of the conflicting mass) into an associative and Markovian rule by storing the previous result of the conjunctive rule and, depending of the rule, other data. Such rules are called quasi-associative and quasi-Markovian.

Some applications require the necessity of **decaying the old sources** because their information is considered to be worn out.

If some bba is not normalized (i.e. the sum of its components is < 1 as in incomplete information, or > 1 as in paraconsistent information) we can easily divide each component by the sum of the components and normalize it. But also it is possible to fusion incomplete and paraconsistent masses, and then normalize them after fusion. Or leave them unnormalized since they are incomplete or paraconsistent.

PCR5 does the most mathematically exact (in the fusion literature) redistribution of the conflicting mass to the elements involved in the conflict, redistribution which exactly follows the tracks of the conjunctive rule.

**Examples of UFT**

Cases:
1. Both sources reliable: use conjunctive rule [default case]:
1.1. A∩B≠ϕ:
   1.1.1. Consensus between A and B; mass → A∩B;
   1.1.2. Neither A∩B nor A∪B interest us; mass → A, B;
1.2. A∩B=ϕ:
    1.2.1. Contradiction between A and B, but optimistic in both of them; mass → A, B;
    1.2.2. One right, one wrong, but don't know which one; mass → AχB;
   1.2.3. Unknown any relation between A and B [default case]; mass → AχB;
   1.2.4. Pessimistic in both A and B; mass → AχB;
   1.2.5. Very pessimistic in both A and B;
   1.2.5.1. Total ignorance ε AχB; mass → AχBχCχD (total ignorance);
   1.2.5.2. Total ignorance = AχB; mass → ϕ (open world);
   1.2.6. A is right, B is wrong; mass → A;
   1.2.7. Both A and B are wrong; mass → C, D;
1.3. Don't know if A∩B = or ≠ ϕ (don't know the exact model); mass → A∩B (keep the mass on intersection till we find out more info) [default case];

2. One source reliable, other not, but not known which one: use disjunctive rule; no normalization needed.
3. S1 reliable, S2 not reliable 20%: discount S2 for 20% and use conjunctive rule.

|                | A    | B    | A∪B  | A∩B  | φ (open world) | A∪B∪C∪D | C   | D   |
|----------------|------|------|------|------|----------------|---------|-----|-----|
| S1             | .2   | .5   | .3   |      |                |         |     |     |
| S2             | .4   | .4   | .2   |      |                |         |     |     |
| S1&S2          | .24  | .42  | .06  | .28  |                |         |     |     |
| S1 or S2       | .08  | .20  | .72  | 0    |                |         |     |     |
| UFT 1.1.1      | .24  | .42  | .06  | .28  |                |         |     |     |
| UFT 1.1.2 (PCR5) | .356 | .584 | .060 | 0  |                |         |     |     |
| UFT 1.2.1      | .356 | .584 | .060 | 0    |                |         |     |     |
| UFT 1.2.2      | .24  | .42  | .34  | 0    |                |         |     |     |
| UFT 1.2.3      | .24  | .42  | .34  | 0    |                |         |     |     |
| UFT 1.2.4      | .24  | .42  | .34  | 0    |                |         |     |     |
| UFT 1.2.5.1    | .24  | .42  | .06  | 0    | 0              | .28     |     |     |
| UFT 1.2.5.2    | .24  | .42  | .06  | 0    | .28            |         |     |     |
| 80% S2         | .32  | .32  | .16  |      |                | .20     |     |     |
| UFT 1.2.6      | .52  | .42  | .06  |      |                |         |     |     |
| UFT 1.2.7      | .24  | .42  | .06  | 0    |                |         | .14 | .14 |
| UFT 1.3        | .24  | .42  | .06  | .28  |                |         |     |     |
| UFT 2          | .08  | .20  | .72  | 0    |                |         |     |     |
| UFT 3          | .232 | .436 | .108 | .224 |                | 0       |     |     |

**Conclusion:**
- A glossary of 32 fusion rules are presented or cited;
- New rules just introduced;
- Most rules first use the conjunctive rule, then transfer the conflicting mass to other sets;
- No rule or fusion theory works in any case;
- Hence, it's needed a <u>Unification of Fusion Theories</u>, which looks like a cooking recipe;
- Default theory of UFT is DSmT, upon the average minimum principle.

**Acknowledgement**.
We want to thank Dr. Wu Li from NASA Langley Research Center, Dr. Philippe Smets from the Université Libre de Bruxelles, Dr. Jean Dezert from ONERA in Paris, Dr.